# Exploiting Heterogeneity in Operational Neural Networks by Synaptic Plasticity

Serkan Kiranyaz[1], Junaid Malik[1,4], Habib Ben Abdallah[1], Turker Ince[2], Alexandros Iosifidis[3] and Moncef Gabbouj[4]

[1] Electrical Engineering, College of Engineering, Qatar University, Qatar;
[2] Electrical & Electronics Engineering Department, Izmir University of Economics, Turkey;
[3] Department of Engineering, Aarhus University, Denmark;
[4] Department of Computing Sciences, Tampere University, Finland.

*Abstract*— **The recently proposed network model, Operational Neural Networks (ONNs), can generalize the conventional Convolutional Neural Networks (CNNs) that are homogenous only with a linear neuron model. As a heterogenous network model, ONNs are based on a generalized neuron model that can encapsulate *any* set of non-linear operators to boost diversity and to learn highly complex and multi-modal functions or spaces with minimal network complexity and training data. However, the default search method to find optimal operators in ONNs, the so-called Greedy Iterative Search (GIS) method, usually takes several training sessions to find a single operator set per layer. This is not only computationally demanding, also the network heterogeneity is limited since the same set of operators will then be used for all neurons in each layer. To address this deficiency and exploit a superior level of heterogeneity, in this study the focus is drawn on searching the best-possible operator set(s) for the hidden neurons of the network based on the "Synaptic Plasticity" paradigm that poses the essential learning theory in biological neurons. During training, each operator set in the library can be evaluated by their synaptic plasticity level, ranked from the worst to the best, and an "elite" ONN can then be configured using the top ranked operator sets found at each hidden layer. Experimental results over highly challenging problems demonstrate that the elite ONNs even with few neurons and layers can achieve a superior learning performance than GIS-based ONNs and as a result the performance gap over the CNNs further widens.**

## I. Introduction

Neurons of a mammalian brain communicate with each other through synaptic connections [1] which control the "strength" of the signals transmitted between neurons via their individual neuro-chemical characteristics. During a learning process, synaptic connections weaken or strengthen according to the amount of stimuli received [2], [3]. This phenomenon is known as "Synaptic Plasticity", which refers to the fact that the connections between nerve cells in the brain are not *static* but can undergo changes, so they are *plastic*. This process is generally accepted to be the major instrument by which living organisms are able to learn [2]. Mammalian brains demonstrate remarkable plasticity, enabling them to alter future behavior, emotions, and responses to sensory input by modifying existing neural circuits [4]-[9]. In other words, during a learning (training) process, the synaptic plasticity implies a significant change (positive or negative) that occurs in the synapse's connection strength as the plastic change often results from the alteration of the number of neurotransmitter receptors located on a synapse. Figure 1 shows a biological neuron and the synaptic connection at the terminal via neurotransmitters. There are several underlying mechanisms that cooperate to achieve the synaptic plasticity, including changes in the quantity of neurotransmitters released into a synapse and changes in how effectively cells respond to those neurotransmitters [5]-[9].

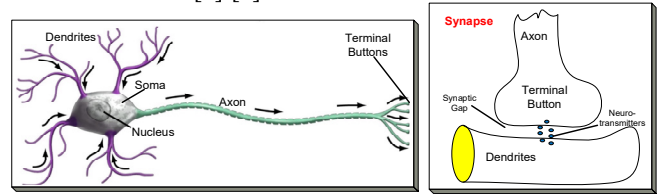

**Figure 1: A biological neuron (left) with the direction of the signal flow and a synapse (right).**

Although the most popular, conventional ANNs such as MLPs and CNNs are designed to mimic biological neurological systems such as the mammalian brain, they can only pose a crude and over-simplistic model due to two primary drawbacks: 1) assumption of a homogenous structure with an identical neuron model across the entire network, 2) use of linear transformation (i.e., linear weighted sum for MLPs or linear convolution for CNNs) [10] as the sole operator. Although non-linear activation functions are employed as a remedy to the latter, the structure still does not match in general to the structural characteristics of the biological neural systems which encapsulate a diverse set of neuron types and varying biochemical and electrophysiological properties [4]-[9]. For example, there are about 55 distinct types of neurons in mammalian retina to realize visual sensing [7]. Most importantly, neurochemical properties of each synaptic connection accomplish the signal operation which is nonlinear in general [11], [12]. Therefore, traditional homogenous ANNs, e.g., Multi-layer Perceptrons (MLPs) [12], [13] with the linear neuron model can only approximate the ongoing learning process that is based on the responses of the training samples, and therefore, they are considered as the "Universal Approximators".

This is perhaps the reason for the significant variations observed in their learning performance. Generally speaking, they are effective when dealing with problems with a monotonous and relatively simpler solution space; however, they may entirely fail when the solution space is highly nonlinear and complex. Although several methods have been proposed in the literature to modify MLPs by changing the network architectures [14]-[21], or the neuron model and/or conventional BP algorithm [23]-[25], or even the parameter updates [26], [27]; the learning performance was not improved significantly, as it is inherently subdued by the underlying homogenous network configuration with the sole linear neuron model.

In order to address this fundamental drawback, recently a heterogeneous and non-linear network model, called Generalized Operational Perceptrons (GOPs) [28], [29], have recently been proposed. GOPs aim at accurately mimicking the actual biological neuron model with *distinct* synaptic connections. In this heterogeneous configuration, the natural diversity that appears in biological neurons and neural networks has been incorporated. Specifically, the diverse set of neurochemical operations in biological neurons (the non-linear synaptic connections plus the integration process occurring in the soma of a biological neuron model) have been modelled by an "operator set" composing of corresponding "Nodal" (for synaptic connection), "Pool" (for integration in soma) and the "Activation" (for activation in the axon) operators.

Therefore, the GOP neuron model naturally becomes a superset of the Perceptrons achieving highly superior performance levels in both function approximation and generalization. The function approximation capability of GOP networks was demonstrated in the studies [28] and [29], where it was shown that GOPs can achieve elegant performance levels on many challenging problems where MLPs entirely fail to learn such as "Two-Spirals", "N-bit Parity" for N>10, "White Noise Regression", etc. The success of GOPs over such problems was unprecedent despite the fact that they were partially heterogenous, i.e., the same operator set is used in all neurons of each layer. Moreover, the generalization capability of GOP networks in classification and ranking problems was shown in [30]-[33], respectively, where an extensive set of experimental comparisons with MLPs has shown that GOP networks can outperform MLPs even when the resulting network structure is much more compact and, thus, computationally more efficient.

Following the example, a heterogenous and non-linear network model, called Operational Neural Network (ONN), has recently been proposed [34] as a superset of CNNs. In this study a similar search method used in [28], [29], called Greedy Iterative Search (GIS), was utilized also in ONNs to find an operator set per layer. The final ONN can then be configured by using the best operator sets found, each of which is assigned to *all* neurons of the corresponding hidden layers. This exhibits several limitations and drawbacks. First and the foremost GIS has a limitation of assigning a single operator set for the entire (hidden) layer neurons due to the two reasons: 1) only the global (network-wise) evaluation is used to "estimate" the HF of the operator set assigned and 2) searching for an individual operator set per neuron prohibits heterogeneity as it creates a search space infeasibly large. Moreover, due to the greedy iterative nature of GIS, to find the best possible operator set for each layer, the number of BP runs required is proportional with the size of the operator set library. As mentioned earlier, the resultant ONN was still a heterogeneous network; however, its heterogeneity was limited as intra-layer homogeneity was still preserved by assigning a single distinct operator set to all neurons within each hidden layer.

In order to adress these limitations and exploit further the heterogenity of ONNs, in this study, we propose a novel configuration approach for Operational Neural Networks (ONNs) based on the synaptic plasticity paradigm. The proposed approach can evaluate the "learning ability" of an individual neuron having a certain operator set in order to measure its contribution to the ongoing learning objective. We claim a direct relation between the tendency of a neuron to alter its connection strengths (synaptic plasticity) and the suitability of its operator set. We quantify this notion by numerically computing the "health factor" of a particular operator assignment to a neuron in a particular hidden layer. Over challenging learning problems and with severe restrictions, it is demonstrated that only an ONN compromised of hidden neurons having an operator set with a high health factor (HF) can achieve the desired learning performance while those with low-HF operator sets fail to contribute well enough and thus result in an inadequate learning performance. Therefore, the fundamental idea is measurement of the HF of each operator set *locally* based on the synaptic plasticity principle. At each hidden layer of the ONN, the HF of each operator set is periodically monitored during a prior BP run and the "promising" operator sets yielding a high HF will gradually be favored to improve the learning performance of the network. The so-called Synaptic Plasticity Monitoring (SPM) sessions will allow the evaluation of each operator set as the network gets mature during the BP training. In order to avoid local minima, the operator set of each neuron will be periodically altered by randomly assigning a new operator set in the library. Since BP is a stochastic process, it may take several sessions to properly evaluate an operator set in a particular layer. The final HF of each operator set will be the average HF reached during the entire BP run with many SPM sessions. Using the final HFs of each operator set in the library for each hidden layer of the network, an "elite" ONN can then be configured using the top operator set(s), each with a density proportional to its HF. Note that the order of the neurons into which an operator set is assigned does not affect the learning performance due to the fully-connected nature of the ONN. The elite ONN configured based on the operators exhibiting the highest synaptic plasticity levels can then be trained to achieve the utmost learning performance. The heterogeneity of the network increases with the number of operator sets used at each layer and on contrary, the least heterogonous ONN can still be configured when only a single operator set (the one with the highest HF ever achieved on that layer) is assigned to all neurons in each layer. In this way, we are able to exploit the role of network heterogeneity over the learning performance evaluated on three challenging problems: image denoising, synthesis and transformation. Finally, with the right sets used in each hidden layer, we shall show that the elite ONN can achieve an superior learning performance can achieve a superior learning performance than GIS configured ONNs and the performance gap over the equivalent CNNs further widens.

The rest of the paper is organized as follows: Section II will briefly present the conventional ONNs while the BP training is summarized in Appendix A (refer to [34] for further details). Section III explains the proposed SPM in detail and shows the configuration of the elite ONN over the final HFs computed for each operator. Section IV first analyzes a sample set of SPM results and then presents a rich set of experiments to exploit the heterogeneity of the network and to perform comparative evaluations between the learning performances of ONNs and CNNs over challenging problems. Finally, Section V concludes the paper and suggests topics for future research.

## II. OPERATIONAL NEURAL NETWORKS

The conventional (deep) 2D CNNs have the classical "linear" neuron model similar in MLPs; however, they further apply two restrictions: kernel-wise (limited) connections and weight sharing. These restrictions turn the linear weighted sum for MLPs to the convolution formula used in CNNs. This is illustrated in Figure 2 (left) where the three consecutive convolutional layers without the sub-sampling (pooling) layers are shown. ONNs borrows the essential idea of GOPs and thus extends the sole usage of linear convolutions in the convolutional neurons by the *nodal* and *pool* operators. This constitute the operational layers and neurons while the two fundamental restrictions, weight sharing and limited (kernel-wise) connectivity, are directly inherited from conventional CNNs. This is also illustrated in Figure 2 (right) where three operational layers and the $k^{th}$ neuron with 3x3 kernels belong to a sample ONN. As illustrated, the input map of the $k^{th}$ neuron at the current layer, $x_k^l$, is obtained by *pooling* the final output maps, $y_i^{l-1}$ of the previous layer neurons *operated* with its corresponding kernels, $w_{ki}^l$, as follows:

$$x_k^l = b_k^l + \sum_{i=1}^{N_{l-1}} oper2D(w_{ki}^l, y_i^{l-1}, 'NoZeroPad')$$

$$x_k^l(m,n)\big|_{(0,0)}^{(M-1,N-1)} = b_k^l + \sum_{i=1}^{N_{l-1}} \left( P_k^l \begin{bmatrix} \Psi_{ki}^l\left(w_{ki}^l(0,0), y_i^{l-1}(m,n)\right), \dots, \\ \Psi_{ki}^l\left(w_{ki}^l(r,t), y_i^{l-1}(m+r,n+t)\right), \dots \end{bmatrix} \right) \quad (1)$$

A close look to Eq. (1) will reveal the fact that when the pool operator is "summation", $P_k^l = \Sigma$, and the nodal operator is "linear", $\Psi_{ki}^l(y_i^{l-1}(m,n), w_{ki}^l(r,t)) = w_{ki}^l(r,t)y_i^{l-1}(m,n)$, for *all* neurons, then the resulting homogenous ONN will be identical to a CNN. Hence, ONNs are indeed the superset of CNNs as the GOPs are the superset of MLPs.

For Back-Propagation (BP) training of an ONN, the following four consecutive stages should be iteratively performed: 1) Computation of the delta error, $\Delta_1^L$, at the output layer, 2) Inter-BP between two consecutive operational layers, 3) Intra-BP in an operational neuron, and 4) Computation of the weight (operator kernel) and bias sensitivities in order to update them at each BP iteration. Stage-3 also takes care of sub-sampling (pooling) operations whenever they are applied in the neuron. BP training is briefly formulated in Appendix A while further details can be obtained from [34].

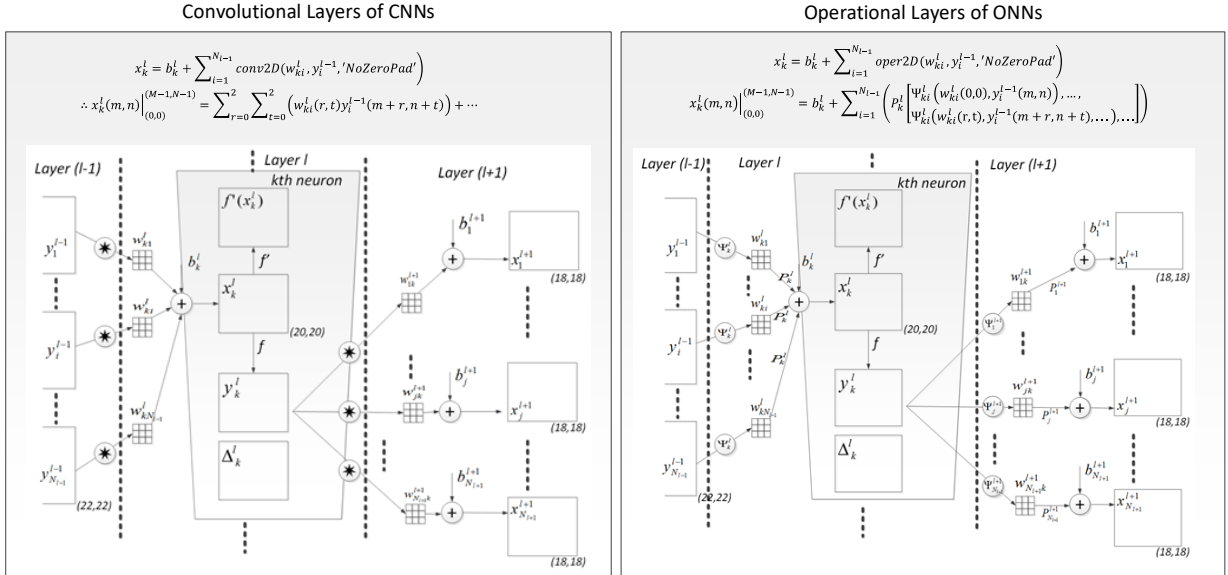

**Figure 2:** Three consecutive convolutional (left) and operational (right) layers with the $k^{th}$ neuron of a CNN (left) and an ONN (right).

## III. SYNAPTIC PLASTICITY MONITORING

Synaptic plasticity is a natural process that enables learning of a new ability or concept so that the brain can (learn to) respond appropriately to the changes in the environment. Therefore, a drastic "variation" of synaptic connection strength indicates an ongoing learning process or equivalently, it confirms the neurons' involvement in the learning process. Conversely, if the neuron's synapses are idle, which means that the strength of the synapses is not *plastic*, rather static, hence the neuron is neither learning nor involved in (or contributing to) the ongoing learning process.

### A. Computation of Health Factors

When this phenomenon is reflected to the neurons of an ONN with varying synaptic connections -or equivalently distinct operators, this can indeed be the right model to measure the health factor (HF) of each hidden neuron when a new operator set is assigned. First of all, the synaptic plasticity of a hidden neuron depends two factors: 1) the hidden layer where it resides, and 2) its operator set. As the prior studies [28], [29] have shown, in a particular hidden layer, neurons with an improper operator set do not contribute to the learning at all, and hence the BP training will fail to achieve the (learning) objective. Therefore, those neurons that contribute the most to the learning process with the best-possible operator set(s) are expected to exhibit a high synaptic plasticity, quantified by a high variation in the connection strength. Therefore, during the synaptic plasticity monitoring (SPM) in the prior BP run, for a certain period of time (number of iterations), the variation in the connection strength of a neuron in a hidden layer can be computed from the strength of its connections (weights) to the next layer neurons.

Consider the case of the $k^{th}$ neuron at layer $l$ with a particular operator set $\theta$ assigned to it. Its output, $y_k^l$, will be utilized by all the neurons in the next layer with their individual weights as expressed in Eq. (10) in Appendix A. Now, the principle of synaptic plasticity entails that if $w_{ik}^{l+1} \forall i \in [1, N_{l+1}]$ undergoes a significant change from its initial value, then the neuron $k$ makes a meaningful contribution to the ongoing learning (training) process. Accordingly, during the prior BP run, once a new operator set, $\theta$, is assigned to the $k^{th}$ neuron at layer $l$, we shall monitor the normalized average strength (power) variation of $w_{ik}^{l+1} \forall i \in [1, N_{l+1}]$ within a sufficient window of BP iterations. Then as expressed in Eq. (2), one can compute the instantaneous health factor $HF_\theta^{k,l}(t)$, by first computing the average weight power (variance), $\bar{\sigma}_k^2(t)$, and then computing the absolute change occurred after a preset number of iterations, M.

$$HF_\theta^{k,l}(t) = \frac{|\bar{\sigma}_k^2(t-M) - \bar{\sigma}_k^2(t)|}{\bar{\sigma}_k^2(t-M)}$$

$$where\ \bar{\sigma}_k^2(t) = \frac{\sum_{i=1}^{N_{l+1}} \sigma_{w_{ik}^{l+1}}^2(t)}{N_{l+1}}\ and \quad (2)$$

$$\sigma_{w_{ik}^{l+1}}^2(t) = \frac{\sum_{r=1}^{K_x}\sum_{p=1}^{K_y}(w_{ik}^{l+1}(r,p) - \mu_w)^2}{K_x K_y}$$

where $\mu_w$ is the sample mean of the weight, $w_{ik}^{l+1}$, of the $k^{th}$ neuron. It is evident that, owing to the stochastic nature of BP, there is no guarantee that the instantaneous health factor, $HF_\theta^{k,l}(t)$ of the operator set, $\theta \in \{\theta_N^*\}$ will correspond to the maximum potential synaptic plasticity level that can be achieved by this particular operator set. Therefore, several (e.g., $S$) SPM sessions will be performed to capture the overall, long-term trend of the synaptic plasticity levels of each operator set. At the end of the prior BP run, for each hidden layer $l$, the final HF of each operator set will be the *average* of all the instantaneous HFs computed during the SPM sessions, as expressed below:

$$HF_\theta^l = \underset{k,t}{\text{avg}}\left(HF_\theta^{k,l}(t)\right) \text{ for } \forall l \in [\![1, L-1]\!] \quad (3)$$

### B. SPM Implementation

SPM is designed to compute the HF of each operator set several times. In this way, the likelihood to "capture" the potential plasticity level of a synaptic (nodal) connection strength (weight variance) is increased. In practice, the monitoring of the strengths (powers) of each hidden neuron's connection weights, $w_{ik}^{l+1}$, is performed periodically at every $M$ iterations. The periodic synaptic plasticity monitoring (SPM) is, therefore, an embedded process into the prior BP that performs three tasks: 1) Computation of the (instantaneous) HF of each operator set at each neuron of each hidden layer, 2) Updating the average HF of each operator set, and 3) Assigning new operator sets *randomly* to each neuron. SPM is designed to improve the network maturity by favoring the "promising" sets with a high HF during random selection in task 3. To accomplish this, once enough statistics are obtained to compute the average HF for each operator set, i.e., after certain number of SPM sessions, $M$, the HFs are unit-normalized to approximate the probability density function (pdf) of the operator sets in the library. The approximate pdf is then used directly in the random operator set assignment in task 3, i.e., probability to choose an operator set is no longer uniform but is the corresponding probability in the pdf, as expressed below"

$$P_l(\theta) = \frac{HF_\theta^l(t)}{\sum_\theta HF_\theta^l(t)}\ for\ \forall l \in [\![1, L-1]\!] \quad (4)$$

In this way the likelihood of choosing an operator is made to be directly proportional to its HF. As a consequence of this improvised random assignment, each random ONN, $ONN^*(\theta)$ that will be configured during later SPM sessions might even have the potential to provide an adequate learning performance on its own, given a sufficient number of BP iterations. In this case one can deduce that the learning problem requires a high level of heterogeneity since $ONN^*(\theta)$ will be highly heterogenous due to the random assignments. Otherwise, the final health factors of the operator sets, $HF_\theta^l$, can then be used to form an elite ONN with the top ranked operator sets at each hidden layer. This will be detailed next.

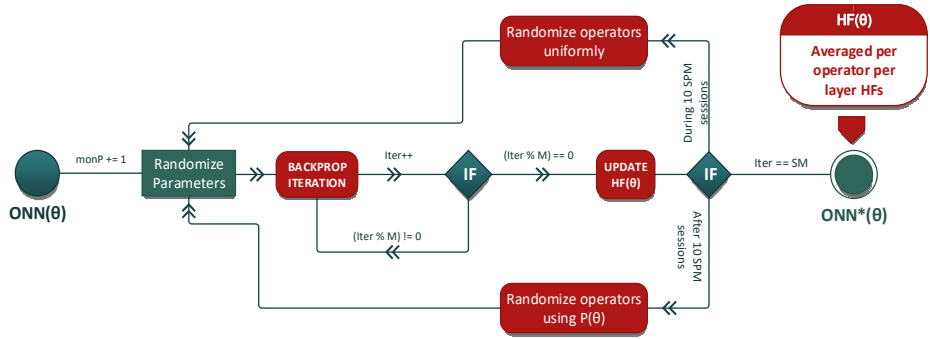

**Figure 3 Flowchart illustrating a single SPM session.**

### C. Configuration of the Elite ONN

The final HFs computed per hidden layer, $HF_\theta^l$, can now be used to form the "elite ONN" for the learning problem in hand. In a particular hidden layer, the order of neurons with a particular operator set does not matter due to fully connected network model, so, the aim is to determine the optimal number of neurons that will have a particular operator set assigned to them. Since the number of hidden neurons in the ONN configuration is fixed in advance, e.g., 12 for the sample ONN used in this study, this is equivalent of finding the optimal "density" for each operator set. Ideally, the density of an operator set should reflect the synaptic plasticity level it has demonstrated during the learning (training) process. Therefore, the density is computed to be proportional to the final HF of that operator set. There can be several approaches among which the heterogeneity level of the ONNs differ. In an extreme case, only the top operator set of a hidden layer with the highest final HF is assigned to all hidden neurons of that layer. This makes "homogenous" layers like the convolutional layers of a conventional CNN, but the ONN network will still be heterogeneous (different operator sets at each layer). However, for certain problems and large ONNs with many neurons such a limited heterogeneity may cause performance degradation. Another extreme case is to use all the operator sets in the library with the densities proportional with HFs; however, this may also cause a practical problem especially for compact ONNs with only few hidden neurons. Ultimately, the question that we seek the answer for is: should an operator set with a relatively low HF be assigned to a neuron or that neuron is used instead for another operator set with higher HF? Especially assigning those operator sets with too low HFs does not make sense in any case. In this study, we shall investigate this for compact ONNs by assigning the *top-S* operator sets having the *S* number of the highest HFs to the neurons of each hidden layer and discard the rest of the sets entirely. One can consider the two extreme cases by simply assigning *S=N* and *S=1*, respectively.

Let $HF_{\theta_i}^l$ for $i = 1:S$, be the final HF of the $i^{th}$ *top-S* operator set at layer $l$. The density, $d_{\theta_i}^l$, and the number of neurons, $n_{\theta_i}^l$ that will be assigned to the $i^{th}$ *top-S* operator set can be expressed as follows:

$$d_{\theta_i}^l = \frac{HF_{\theta_i}^l}{\sum_{i=1}^{S} HF_{\theta_i}^l} \rightarrow n_{\theta_i}^l = \lfloor N_l d_{\theta_i}^l \rfloor \text{ for i=2:S}$$

$$n_{\theta_1}^l = N_l - \sum_{i=2}^{S} n_{\theta_i}^l$$

(5)

where $N^l$ is the total number of hidden neurons at layer $l$ and $\lfloor . \rfloor$ denotes the *floor* operator. Note that the number of neurons to which the operator set with the highest HF is assigned, $n_{\theta_1}^l$, is computed after all the other sets are assigned. This will ensure that all the neurons at layer $l$ will have an operator set assigned.

IV. EXPERIMENTAL RESULTS

In this section we perform comparative evaluations of the elite ONNs, configured by SPM as presented in Section III.C, over three challenging problems: 1) Image Denoising, 2) Image Syntheses, and 3) Image Transformation. The latter two problems are common with [34] while we tried ONNs over a more challenging Image Denoising problem. In order to demonstrate the learning capabilities of the elite ONNs better, we introduce the same training constraints:

i) Low Resolution: We keep the image resolution very low, e.g., thumbnail size (i.e., 60x60 pixels).
ii) Compact Model: We keep the elite ONN configuration compact, e.g., only two hidden layers with 24 hidden neurons, i.e., *Inx12x12xOut* as shown in Figure 4.
iii) Scarce Training Data: For problems that require learning a generalized model such as image denoising, we train the network over a limited data (i.e., only 10% of the dataset) while testing over the rest with a 10-fold cross validation.
iv) Multiple Regressions: For the two regression problems (image syntheses and transformation), a single network is trained to regress multiple (e.g., 4-8) images at once.

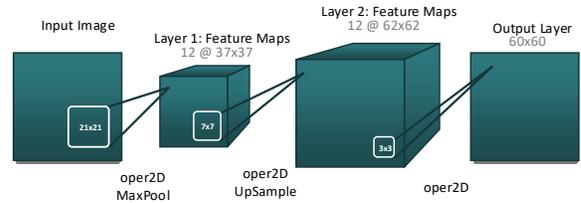

**Figure 4 Architecture of the compact neural network architecture used for experiments in this study.**

*A. Experimental Setup*

In any BP training session, for each iteration, *t*, with the MSE obtained at the output layer, $E(t)$, a global adaptation of the learning rate, $\varepsilon$, is performed within the range [$2.10^{-1}$, $1.10^{-6}$], as follows:

$$\varepsilon(t) = \begin{cases} \alpha\varepsilon(t-1) & \text{if } E(t) < E(t-1) \text{ and } \alpha\varepsilon(t-1) \leq 5.10^{-1} \\ \beta\varepsilon(t-1) & \text{if } E(t) \geq E(t-1) \text{ and } \beta\varepsilon(t-1) \geq 5.10^{-5} \\ \varepsilon(t-1) & \text{else} \end{cases}$$

(6)

where α=1.05 and β=0.7, respectively. Since BP training is based on stochastic gradient descent, for each problem we shall perform 10 BP runs, each with random parameter initialization.

**Table 1: Nodal operators and derivatives**

| *i* | Function | $\Psi_i^{l+1}(y_k^l, w_{ik}^{l+1})$ | $\nabla_w \Psi_{ki}^{l+1}$ | $\nabla_y \Psi_{ki}^{l+1}$ |
|---|---|---|---|---|
| 0 | Linear | $w_{ik}^{l+1} y_k^l$ | $y_k^l$ | $w_{ik}^{l+1}$ |
| 1 | Cubic | $K w_{ik}^{l+1} (y_k^l)^3$ | $K(y_k^l)^3$ | $3K w_{ik}^{l+1} (y_k^l)^2$ |
| 2 | Sine | $\sin(K w_{ik}^{l+1} y_k^l)$ | $K y_k^l \cos(K w_{ik}^{l+1} y_k^l)$ | $K w_{ik}^{l+1} \cos(K w_{ik}^{l+1} y_k^l)$ |
| 3 | Exp. | $\exp(w_{ik}^{l+1} y_k^l) - 1$ | $y_k^l \exp(w_{ik}^{l+1} y_k^l)$ | $w_{ik}^{l+1} \exp(w_{ik}^{l+1} y_k^l)$ |
| 4 | Sinh | $\sinh(K w_{ik}^{l+1} y_k^l)$ | $K y_k^l \cosh(K w_{ik}^{l+1} y_k^l)$ | $K w_{ik}^{l+1} \cosh(K w_{ik}^{l+1} y_k^l)$ |
| 5 | Sinc | $\sin(K w_{ik}^{l+1} y_k^l) / y_k^l$ | $K \cos(K w_{ik}^{l+1} y_k^l)$ | $(K w_{ik}^{l+1} \cos(K w_{ik}^{l+1} y_k^l) / y_k^l) - (\sin(K w_{ik}^{l+1} y_k^l) / (y_k^l)^2)$ |
| 6 | Chirp | $\sin(K_C w_{ik}^{l+1} (y_k^l)^2)$ | $K_C (y_k^l)^2 \cos(K w_{ik}^{l+1} (y_k^l)^2)$ | $2 K_C w_{ik}^{l+1} y_k^l \cos(K_C w_{ik}^{l+1} (y_k^l)^2)$ |

### Table 2: Pool operators and derivatives

| $i$ | Function | $P_i^{l+1}[..,\Psi_i^{l+1}(y_k^l, w_{ik}^{l+1}),..]$ | $\nabla_{\Psi_{ki}} P_i^{l+1}$ |
|---|---|---|---|
| 0 | Summation | $\sum_{k=1}^{N_l} \Psi_i^{l+1}(w_{ik}^{l+1}, y_k^l)$ | 1 |
| 1 | Median | $\underset{k}{median}(\Psi_i^{l+1}(w_{ik}^{l+1}, y_k^l))$ | $\begin{cases} 1 \text{ if arg } median(\Psi_i^{l+1}(w_{ik}^{l+1}, y_k^l)) = k \\ 0 \qquad\qquad else \end{cases}$ |

### Table 3: Activation operators and derivatives

| $i$ | Function | $f(x)$ | $f'(x)$ |
|---|---|---|---|
| 0 | Tangent hyperbolic | $tanh(x) = \dfrac{1 - e^{-2x}}{1 + e^{-2x}}$ | $1 - f(x)^2$ |
| 1 | Linear-Cut | $lin\text{-}cut(x) = \begin{cases} x/cut & if\ |x| \leq cut \\ -1 & if\ x < -cut \\ 1 & if\ x > cut \end{cases}$ | $\begin{cases} 1/cut & if\ |x| \leq cut \\ 0 & else \end{cases}$ |

The operator set library that is used to form the ONNs to tackle the challenging learning problems in this study is composed of a few essential nodal, pool and activation operators. Table 1 presents the 7 nodal operators used along with their partial derivatives, $\nabla_w \Psi_{ki}^{l+1}$ and $\nabla_y \Psi_{ki}^{l+1}$ with respect to the weight, $w_{ik}^{l+1}$, and the output, $y_k^l$ of the previous layer neuron respectively. Similarly, Table 2 presents the two common pool operators and their derivatives with respect to the nodal term, $\sum_{k=1}^{N_l} \Psi_i^{l+1}(w_{ik}^{l+1}, y_k^l)$. Finally, Table 3 presents the two common activation functions (operators) and their derivatives (*cut* = 10 for the *lin-cut* operator). Using these lookup tables, the error at the output layer can be backpropagated and the weight sensitivities can be computed. The top section of Table 4 enumerates each potential operator set and the bottom section presents the index of each individual operator set in the operator set library, $\{\theta_N^*\}$, which will be used in all experiments. There is a total of $N=7\times2\times2=28$ distinct operator sets that constitute the operator set library, $\{\theta_N^*\}$. Let $\theta_i: \{i_{pool}, i_{act}, i_{nodal}\}$ be the $i^{th}$ operator set in the library. Note that the first operator set, $\theta_0: \{0,0,0\}$ with index $i = 0$, belongs to the native operators of a CNN to perform linear convolution with traditional activation function, *tanh*. In accordance with the activation operators used, the dynamic ranges of the input/output images in all problems are normalized to within [-1, 1] as follows:

$$I_i = 2\dfrac{I_i - \min(I)}{\max(I) - \min(I)} - 1 \qquad (7)$$

where $I_i$ is the $i^{th}$ pixel value in an image, $I$.

As mentioned earlier, and illustrated in Figure 4, the same compact network configuration with only two hidden layers and a total of 24 hidden neurons, *Inx12x12xOut* is used in all the experiments. The first hidden layer applies sub-sampling by $ssx = ssy = 2$, and the second one applies up-sampling by $usx = usy = 2$.

### Table 4: Operator enumeration (top) and the index of each operator set (bottom).

| $i$ | 0 | 1 | 2 | 3 | 4 | 5 | 6 |
|---|---|---|---|---|---|---|---|
| Pool | sum | median | | | | | |
| Act. | tanh | lin-cut | | | | | |
| Nodal | mul. | cubic | sin | exp | sinh | sinc | chirp |

| $\{\theta_N^*\}$ Index | Pool | Act. | Nodal |
|---|---|---|---|
| 0 | 0 | 0 | 0 |
| 1 | 0 | 0 | 1 |
| 2 | 0 | 0 | 2 |
| 3 | 0 | 0 | 3 |
| 4 | 0 | 0 | 4 |
| 5 | 0 | 0 | 5 |
| 6 | 0 | 0 | 6 |
| 7 | 0 | 1 | 0 |
| 8 | 0 | 1 | 1 |
| … | … | … | … |
| 26 | 1 | 1 | 5 |
| 27 | 1 | 1 | 6 |

### B. SPM Results

During each SPM session, average weight power (variance), $\bar{\sigma}_k^2(t)$ and health factors for operator sets, $HF_\theta^{k,l}(t)$, assigned to hidden neurons ($k=1:12$) of each hidden layer ($l=1:2$) are computed. Figure 5 shows the average weight power (variance), $\bar{\sigma}_k^2(t)$ plots of some operator sets which belong to the hidden neurons at the 2nd hidden layer, during the first SPM session of the prior BP run. The operator search is performed for the 1st fold of the Image Transformation problem where a single network learns to perform four distinct image-to-image translations. . The duration of the monitoring window $M$ is kept as 80 iterations, starting at the first iteration. Except the first problem (denoising) we omit the usage of the "median" pool operator, and therefore, reduce the operator set library's cardinality to 1x2x7=14. It is apparent from Figure 6 that the operator set with the highest and lowest $HF_\theta^2$ is $\theta = 6: (0, 0, 6)$ for the pool (*sum*=0), activation (*tanh*=0) and nodal (*chirp*=6) and $\theta = 0: (0, 0, 0)$ for the native CNN operator, linear convolution, respectively. Note that during the random operator assignment to layer 2, the operator set $\theta = 4: (0, 0, 4)$ with nodal operator *Hyperbolic Sine* (*sinh*) is assigned to 4 neurons and hence there are 4 plots of $\bar{\sigma}_k^2(t)$ each corresponding to a different, hidden neuron. For this set, maximum health factor, i.e., $\max_k HF^{k,2}_{\theta=4}(80)$, computed among the 4 neurons is presented in the plot. The omitted plots belong to the operator sets with indices 7 ($HF^{k,2}_{\theta=7}(80) = 0.01$), 8 ($HF^{k,2}_{\theta=8}(80) = 0.03$) and 12 ($HF^{k,2}_{\theta=12}(80) = 0.05$). Note that these operator sets share the same nodal and pool operators with the sets, 0, 1 and 5, respectively. Obviously the 2$^{nd}$ activation function (lin-cut) has failed to improve the HF for the operator sets 0 and 1 while much worse HF ($HF^{k,2}_{\theta=12}(80) = 0.05$) is obtained compared to $HF^{k,2}_{\theta=5}(80) = 0.80$. However, it is too early to make any decisive judgements on the learning capability of each operator set with a single SPM session over an immature ONN. As reasoned earlier, several SPM sessions are indeed required within the prior BP run to accurately approximate the true synaptic plasticity level of each operator set.

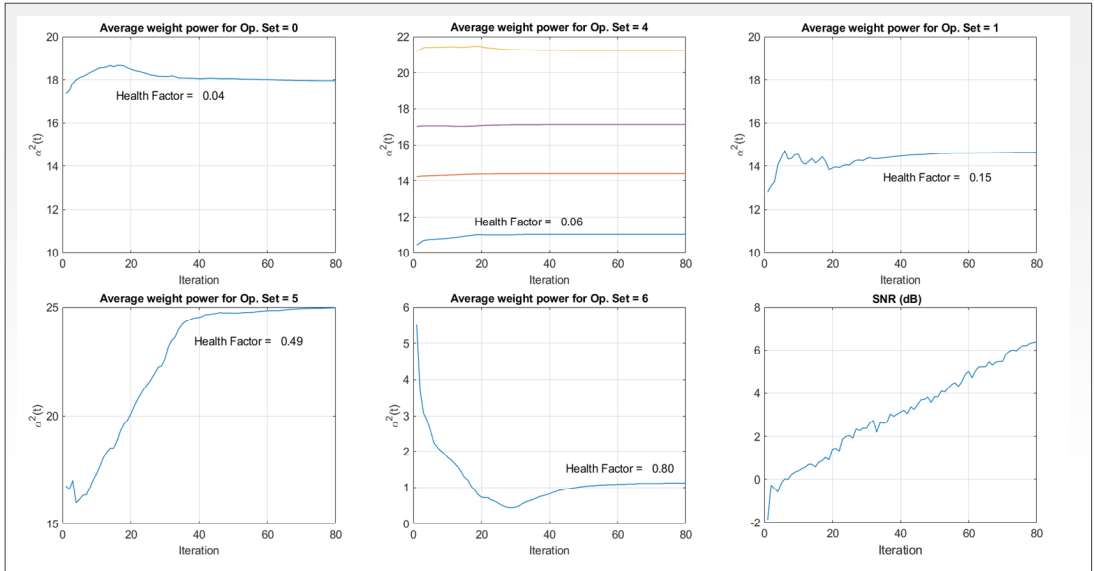

**Figure 5:** For Image Transformation problem (1$^{st}$ cross-validation), SNR (bottom, right) and average weight power (variance), $\bar{\sigma}^2_k(t)$ plots for 2$^{nd}$ layer neurons with operator sets, $\theta$: 0, 1, 4, 5 and 6. This is for the first (p=1) SPM session of the prior BP of the sample ONN.

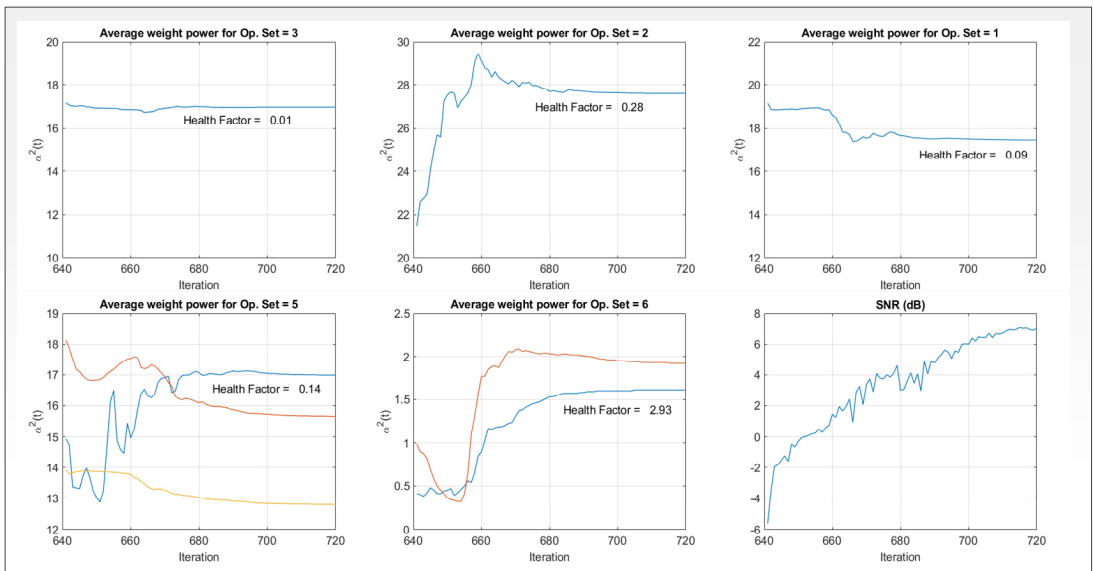

**Figure 6:** For Image Transformation problem (1$^{st}$ cross-validation), SNR (bottom, right) and average weight power (variance), $\bar{\sigma}^2_k(t)$ plots for 2$^{nd}$ layer neurons with operator sets, $\theta$: 1, 2, 3, 5 and 6. This is for the 8$^{th}$ (p=8) SPM session of the prior BP of the sample ONN.

Figure 6 shows the same plots of some operator for the 8$^{th}$ SPM session starting at BP iteration 640. The omitted plots belong to the operator sets with indices 12 ($HF^{k,2}_{\theta=12}(720) = 0.16$) and 13 ($HF^{k,2}_{\theta=13}(720) = 0.38$). Note that these operator sets share the same nodal and pool operators with the sets, 5 and 6, respectively; however, their learning performance is, so far, worse due to the different activation operator used. Comparing with the corresponding plots in Figure 5, several observations can be made. First of all, plots for operator sets 0 and 4 in Figure 5 no longer exist since they are among the worst operator sets for layer 2, their neurons are considered as unhealthy and thus new sets in the SPM pool will randomly be assigned. On the other hand, there are two new operator sets, 2 and 3, that were selected into the SPM pool before. Among the common operator sets, 1, 5 and 6, $HF_{\theta=1}^{k,2}(720) = 0.09$ and especially $HF_{\theta=5}^{k,2}(720) = 0.14$ are now much lower than the corresponding health factors at the first SPM session, while the $HF_{\theta=6}^{k,2}(720)=2.93$ becomes significantly higher than before.

Table 5 presents the final health factors, $HF_\theta^l$, at the end of the three prior BP runs. For layer 2, the best operator set is $\theta = 6$ with a significantly high HF. The second-best set is $\theta$ =13, which uses the same pool (*sum*) and nodal operator (*chirp*) but a different activation function (*lin-cut*). However, it has achieved much lower HF than $\theta = 6$, while the third best set is $\theta =2$. All the other operator sets have achieved HFs 0.2 or below. For layer 1, operator set $\theta$ =5 has the highest HF but, unlike layer 2, the margin is quite slim. This basically shows that several operator sets can be used for the first hidden layer. The set $\theta = 11$ has the lowest HF for both hidden layers and thus became the worst operator set for this problem.

**Table 5:** For Image Transformation problem (1$^{st}$ cross-validation), at the end of the prior BP run for 1$^{st}$ X-validation set, the final health factors, $HF_\theta^l$, are presented below for all operator sets in the library, $\forall \theta \in \{\theta_N^*\}$, assigned to neurons in levels 1 and 2 (L1 and L2). The maximum (bold) and minimum (red) HF values are highlighted.

| $\theta$: | 0 | 1 | 2 | 3 | 4 | 5 | 6 | 7 | 8 | 9 | 10 | 11 | 12 | 13 |
|---|---|---|---|---|---|---|---|---|---|---|---|---|---|---|
| L1 | 1.15 | 1.15 | 0.97 | 1.26 | 0.82 | **1.31** | 0.74 | 0.85 | 1.09 | 0.88 | 0.74 | 0.33 | 1.12 | 0.80 |
| L2 | 0.09 | 0.22 | 0.20 | 0.18 | 0.02 | 0.13 | **0.67** | 0.04 | 0.20 | 0.17 | 0.04 | 0.00 | 0.17 | 0.23 |

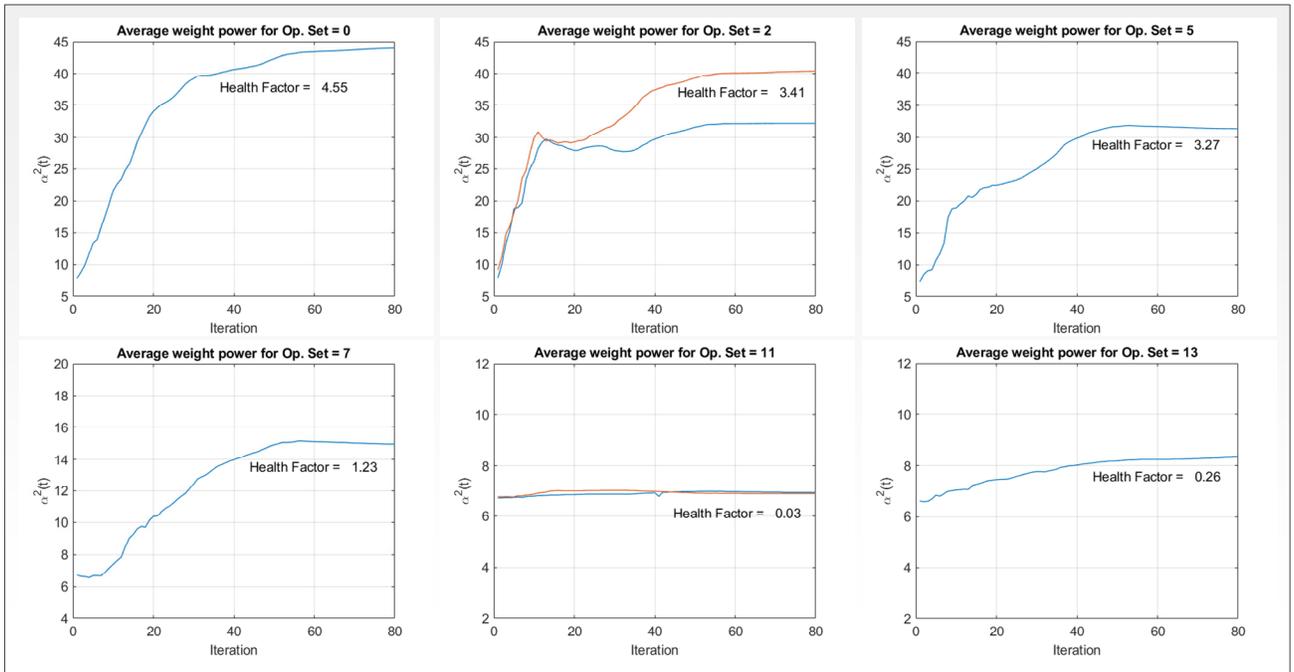

**Figure 7:** For Image Transformation problem (3$^{rd}$ cross-validation), the average weight power (variance), $\bar{\sigma}_k^2(t)$ plots for 1$^{st}$ layer neurons with operator sets, $\theta$: 0, 2, 5, 7, 11 and 13. This is for the 1$^{st}$ (p=1) SPM session of the prior BP of the sample ONN.

**Table 6:** For Image Transformation problem (3$^{rd}$ cross-validation), at the end of the prior BP run, the final health factors, $HF_\theta^l$, are presented below for all operator sets in the library, $\forall \theta \in \{\theta_N^*\}$, assigned to neurons in levels 1 and 2 (L1 and L2). The maximum (bold) and minimum (red) HF values are highlighted.

| $\theta$: | 0 | 1 | 2 | 3 | 4 | 5 | 6 | 7 | 8 | 9 | 10 | 11 | 12 | 13 |
|---|---|---|---|---|---|---|---|---|---|---|---|---|---|---|
| L1 | 2.03 | 1.70 | 1.37 | 1.24 | 1.59 | 1.39 | 0.90 | 1.80 | **2.10** | 1.79 | 2.01 | 0.39 | 1.72 | 0.53 |
| L2 | 0.13 | 0.12 | 0.16 | 0.13 | 0.06 | 0.24 | **0.72** | 0.16 | 0.20 | 0.26 | 0.10 | 0.01 | 0.13 | 0.24 |

On another Image Transformation fold (3$^{rd}$), Figure 7 shows the average weight power (variance), $\bar{\sigma}_k^2(t)$ plots of some operator sets during the first SPM session of the prior BP run. This time the operator set 0 (linear convolution) has achieved the highest HF among others. At the end of the prior BP run, the final HFs, $HF_\theta^l$, presented in Table 6 indicates the top three operator sets for layer 1 are 8, 0 and 10. One can also notice that the operator set $\theta$ =6 has consistently the highest HF for layer 2, and the set $\theta = 11$ has again the lowest HFs for both layers.

## C. Comparative Evaluations and Results

In order to evaluate the learning performance of the ONNs for the three regression problems, image denoising, synthesis and transformation, we used the Signal-to-Noise Ratio (SNR) evaluation metric, which is defined as the ratio of the signal power to noise power, i.e., $SNR = 10\log(\sigma_{signal}^2/\sigma_{noise}^2)$. The ground-truth image is the original signal and its difference to the actual output yields the "noise" image. Besides the comparative evaluations of each problem tackled by the elite ONNs and conventional CNNs, the following sub-sections will especially present a detailed evaluation of the proposed SPM for finding the best and also the worst (set of) operator set(s) for each problem. Finally, an experimental analysis will exploit the role of heterogeneity in ONNs to achieve the maximum learning performance. For this we shall evaluate the best ONN trained during the prior BP, ONN*($\theta$), and the ONNs configured by *top-S* ranked operator sets used and then trained from scratch with 10 individual BP runs among which the best ONN with the highest learning performance will be taken. Similarly, for a fair comparison, 10 individual BP runs, each compromising of 240 iterations, are performed and the best network with the top performance on the training set is selected for the comparative evaluation.

### 1) Image Denoising

The denoising application for additive White Gaussian Noise (AWGN) is a typical domain for deep CNNs that have recently achieved state-of-the-art performances [35]-[39]. ONNs outperformed CNNs over AWGN denoising [34] with a SNR gap around 1 dB. In order to make a more challenging application, in this study we corrupted images by "Salt and Pepper" noise with *p*=0.4 probability. Unlike traditional approaches, which mostly deal with mild noise, such a noise probability is so high that makes most of the corrupted images practically incomprehensible by the naked eye. In order to perform comparative evaluations, we used 1000 images from Pascal VOC database. Such a restriction has been applied in order to evaluate the learning potential of ONNs for harsh denoising applications. The dataset is partitioned into train (10%) and test (90%) with 10-fold cross validation. So, for each fold, both network types are trained 10 times by BP over the train (100 images) partition and tested over the rest (900 images). For this problem, we have used the pool operator, "*median*" but omit the activation operator "*lin-cut*". Therefore, the operator library is still composed of 2x1x7=14 operator sets. The prior BP run for SPM has been performed only *once* in order to rank the operator sets per layer, and then an elite ONN formed with *top-S* ranked sets is then trained for each fold. At the end, the average performances (over both train and test partitions) of the 10-fold cross validation are compared for the final evaluation.

Figure 8 shows the HF bar plots per operator set from the prior BP run of the first fold. For layer-1, the top-4 operator sets are $\theta = 2, \theta = 8, \theta = 0$ and $\theta = 10$. It is expected to see the two top operators ($\theta = 8\ and\ 10$) use *median* pool operator in the first layer. For layer-2, the set $\theta = 4$, with nodal operator *sinh* becomes the best with a significant margin. The 2$^{nd}$ and 3$^{rd}$ top operators are $\theta = 3$ and $\theta = 0$ with *exponential* and *linear* nodal operators, respectively.

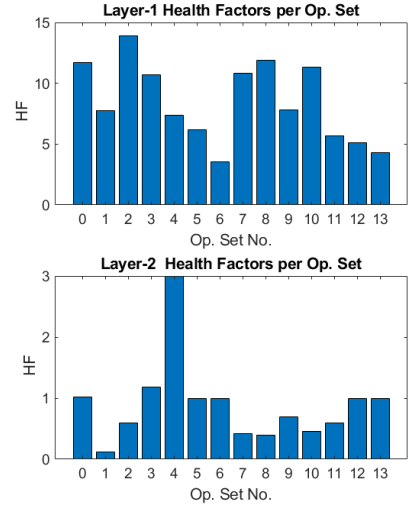

**Figure 8:** $HF_\theta^l$ vs. $\theta$ plots for *l*=1 (top) and *l*=2 (bottom) used in all 10-fold cross validation runs for Image Denoising.

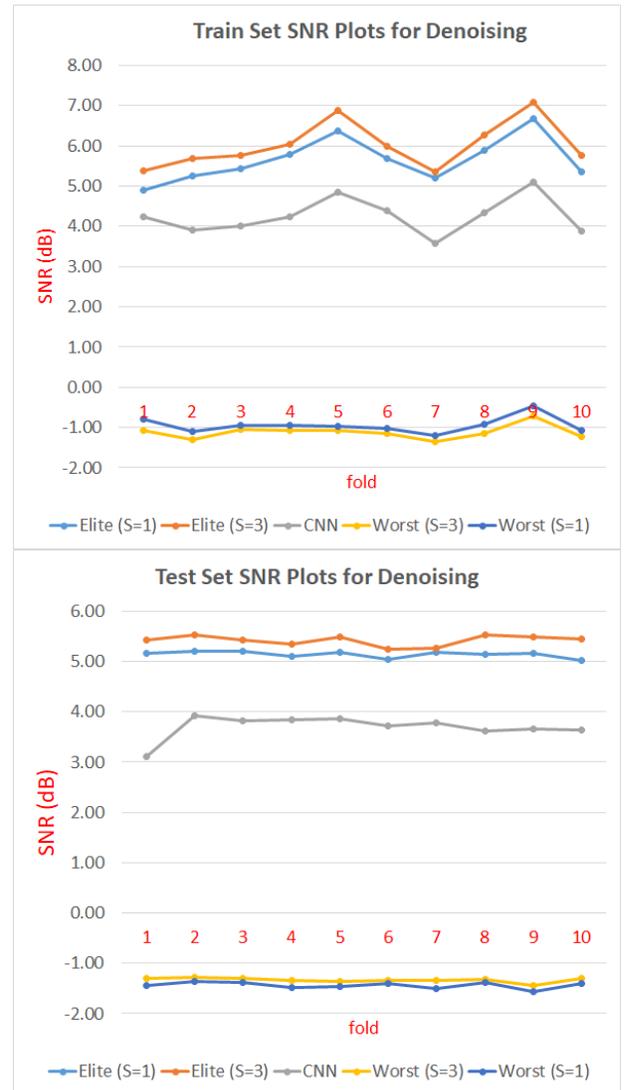

**Figure 9:** Best SNR levels for each denoising fold achieved by the top-performing CNNs (grey) and elite ONNs for the train (top) and test (bottom) set partitions of Pascal database.

**Table 7: Train and test average SNRs achieved in Image Denoising during 10-fold cross-validation. The best (bold) and worst (red) SNR values are highlighted.**

| Avg. SNR | Elite (S=1) | Elite (S=3) | CNN | Worst (S=3) | Worst (S=1) |
|---|---|---|---|---|---|
| Train | 5.74 | **6.09** | 4.25 | -1.12 | -0.96 |
| Test | 5.14 | **5.42** | 3.70 | -1.34 | -1.44 |

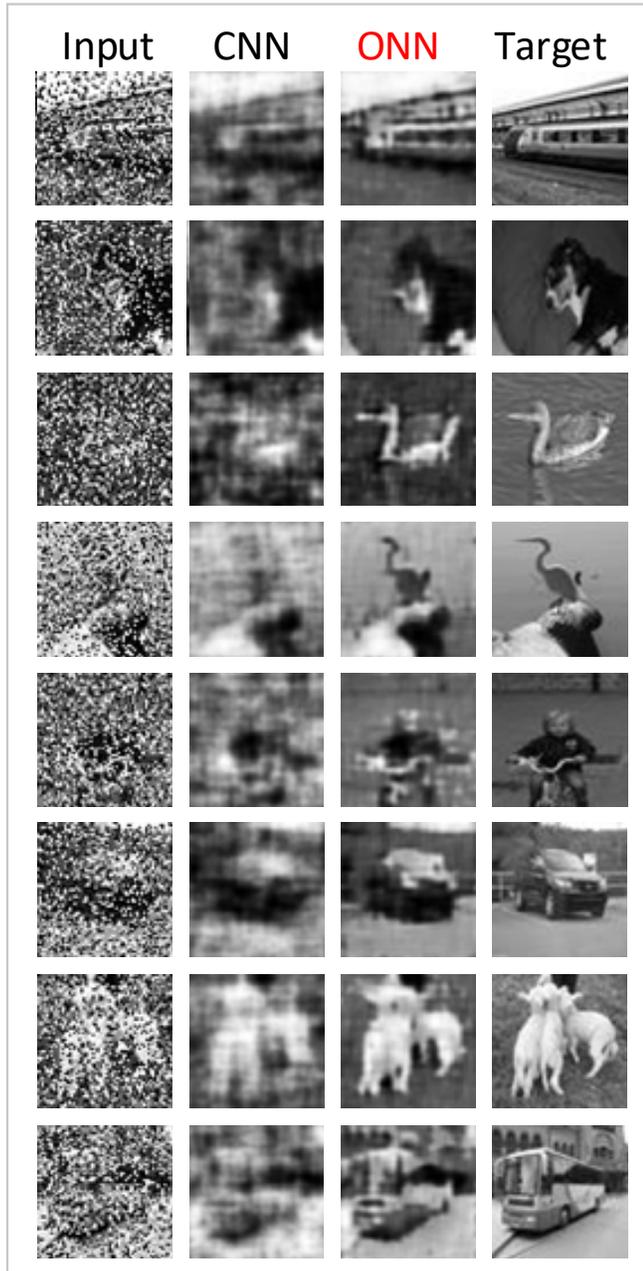

**Figure 10:** Randomly selected original (target) and noisy (input) images and the corresponding outputs of the best CNNs and ONNs from the train partition.

Interestingly, the native sole operator of a CNN, $\theta = 0$ (convolution) has the 3$^{rd}$ rank in both layers. Especially, for layer-1, the difference between $HF_0^1$ and $HF_2^1$ or $HF_8^1$ is not significant. However, for layer-2 there is a significant gap between $HF_0^2$ and the HF of the top operator set, $HF_4^2$. Therefore, one can expect that this will naturally reflect on the performances of the CNN and the elite ONN.

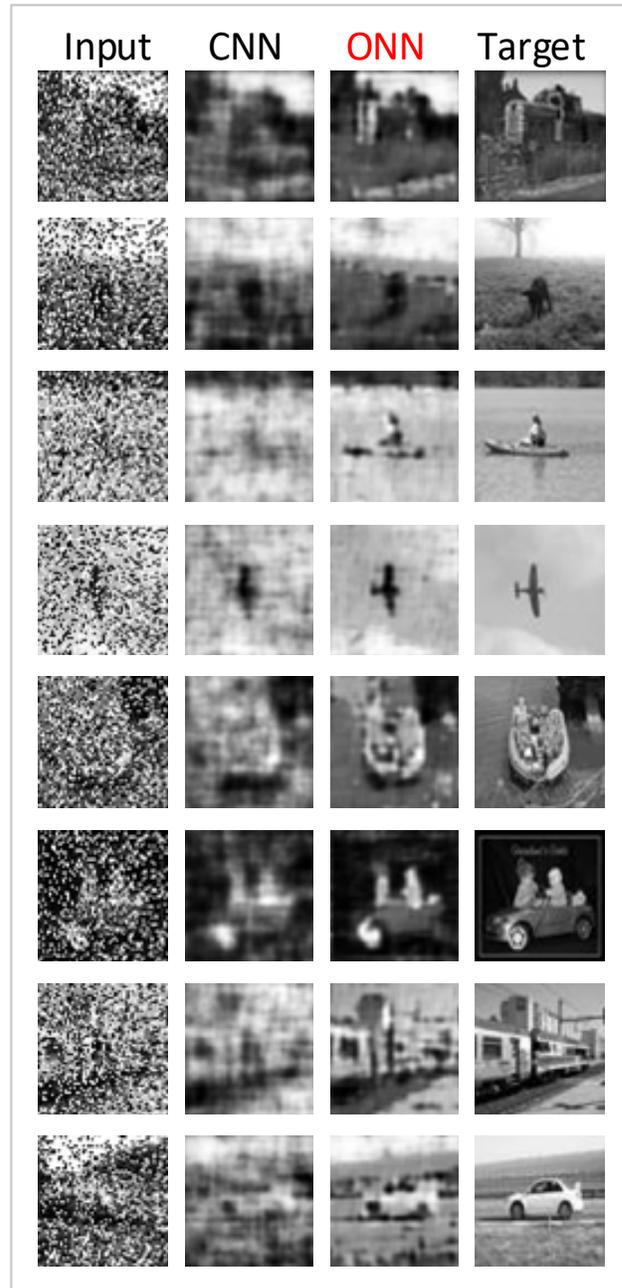

**Figure 11:** Randomly selected original (target) and noisy (input) images and the corresponding outputs of the best CNNs and ONNs from the test partition.

Figure 9 shows the train and test SNR plots of the top-performing CNNs and ONNs among the 10 BP runs corresponding to each fold whereas Table 7 presents the average SNR values, respectively. The results clearly show that the elite and the worst ONNs configured according to the *top-S* and *bottom-S* operator sets found during SPM obtain the best and the worst results. This basically validates the notion that an SPM during a prior BP run can indeed rank the operator sets at each layer accurately from the best to the worst. For this problem, it is quite apparent that the elite ONN with S=3 performs significantly better than the one with S=1. This shows that increasing the heterogeneity level of the ONN improves both the learning and generalization performance for image denoising. The elite ONN with *S*=3 significantly surpasses the CNN (> 1.5 dB on both train and test average SNR values).

For a visual evaluation, Figure 10 and Figure 11 show randomly selected original (target),noisy (input) images and the corresponding outputs of the best CNNs and ONNs from the test partition. Over both train and test denoising examples, the elite ONN exhibits a substantial increase in denoising performance from those highly corrupted input images. Most of the object edges are intact and the background uniformity has been highly preserved. On the other hand, the CNN results show a severe blurring on the edges and deterioration of both foreground and background textual patterns, some of which makes it impossible to realize the true content of the restored image.

*2) Image Synthesis*

Image syntheses is a typical regression problem where a single network learns to synthesize a set of images from individual noise (WGN) images. As recommended in [34], we train a single network to (learn to) synthesize 8 (target) images from 8 WGN (input) images, as illustrated in Figure 12. We repeat the experiment 10 times (folds) each with different set of target images, so 8x10=80 images randomly selected from Pascal VOC dataset. The gray-scaled and down-sampled original images are the target outputs while the WGN images are the input. For each trial, we first performed a single prior BP run with *S*=30 SPM sessions each with M=80 iterations in order to compute $HF_\theta^l$ using which 2 elite (and 2 worst) ONNs are configured with *top-S* and *bottom-S* operator sets for *S*=1 and *S*=3.

Figure 13 shows the average HF bar plots per operator set from each prior BP run of the 10-fold cross validation. For layer-1, the top three operator sets are usually $\theta = 10$ and $\theta = 9 \; or \; 8$, and sometimes $\theta = 7$. It appears that the activation operator (lin-cut) makes a significant difference in the top operator sets with nodal operators *exp*, *cubic* and *sine*. The operator set, $\theta = 7$, corresponds to linear convolution with lin-cut activation operator, and in 3 out of 10 cross-validations, it was one of the top-3 operator sets. For layer-2, the set $\theta = 6$, with nodal operator *chirp* was always the top operator set in all cross validation runs without any exception. It is indeed the most dominant set almost in all runs with the highest final HF usually more than 3 times higher than any other. The 2[nd] and 3[rd] top operators were usually $\theta = 9$ and $\theta = 13$ with *sine* and *chirp* nodal operators, respectively. On the other hand, the worst three operators for this layer are $\theta = 4 \; or \; 11$, $\theta = 3 \; or \; 10$, and $\theta = 0 \; or \; 7$, which correspond to *sinh, exp* and *linear convolution* operations respectively.

Figure 14 shows the SNR plots of the top-performing CNNs and elite ONNs among the 10 BP runs for each syntheses experiment (fold) whereas Table 8 presents the average SNR and MSE levels, respectively. The results again show that the elite and the worst ONNs configured according to the *top-S* and *bottom-S* operator sets perform as expected. For this particular problem, it is quite apparent that the elite ONN with S=1 performs significantly better than the one with S=3. This is actually an expected outcome due to the superiority of the operator set, $\theta = 6$, whereas the usage of the 2[nd] and 3[rd] top operator sets instead of the best operator set degrades the learning performance. In other words, for this problem increasing the heterogeneity level of the ONN does not improve the learning performance. In terms of average SNR, the elite ONN with *S*=1 significantly surpasses the CNN (> 2.5 dB) and also surpassed the GIS-configured ONN in [34] around 1.3 dB despite the fact that the total number of hidden neurons is less (24 vs. 48). This basically shows that SPM can find better operator sets than GIS.

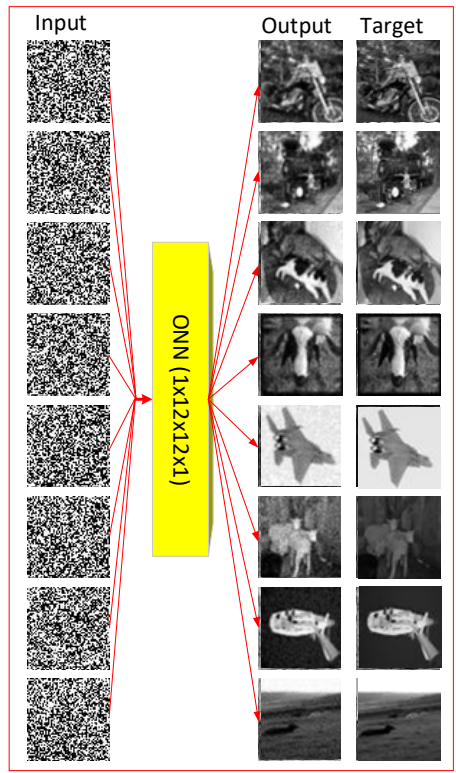

**Figure 12: The outputs of the BP-trained ONN with the corresponding input (WGN) and target (original) images from the 2[nd] syntheses fold.**

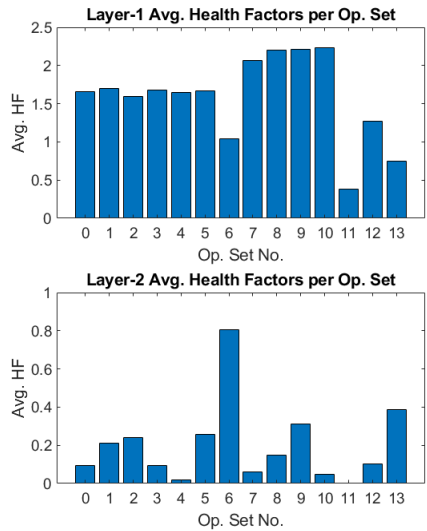

**Figure 13: Average $HF_\theta^l$ vs. $\theta$ plots for *l*=1 (top) and *l*=2 (bottom) from 10-fold X-validation runs for Image Synthesis.**

For a visual comparative evaluation, Figure 15 shows a random set of 14 syntheses outputs of the best CNNs and elite ONNs with the target image. The performance gap is also clear here especially some of the CNN outputs have suffered from severe blurring and/or textural artefacts.

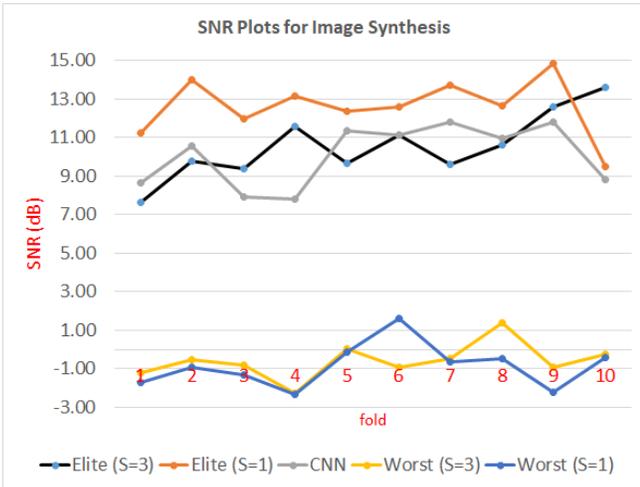

**Figure 14:** Best SNR levels for each synthesis fold achieved by the top-performing CNNs (grey) and ONNs.

**Table 8:** Average SNR and MSE achieved in Image Synthesis during 10-fold cross-validation. The best (bold) and worst (red) SNR values are highlighted.

| Avg. | Elite (S=1) | Elite (S=3) | CNN | Worst (S=3) | Worst (S=1) |
|---|---|---|---|---|---|
| SNR | **12.76** | 10.88 | 10.23 | -0.55 | -0.76 |
| MSE | **0.77** | 1.24 | 1.34 | 18.64 | 20.91 |

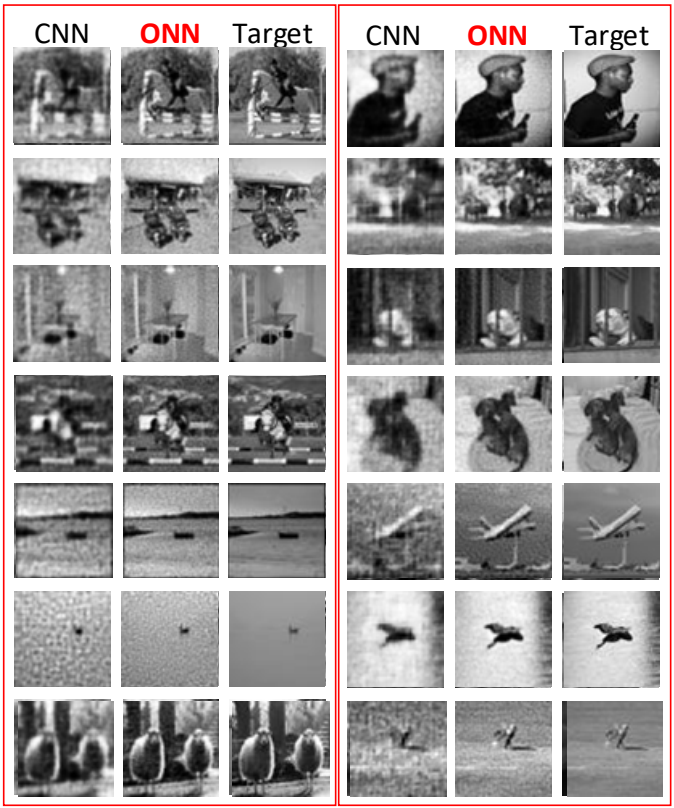

**Figure 15:** A random set of 14 synthesis outputs of the best CNNs and ONNs with the target images. The WGN input images are omitted.

### 3) Image Transformation

This problem aims to test whether a network can (learn to) transform one set of images to another. It is commonly referred to as "Image-to-Image Translation" in contemporary literature and various Deep CNNs have recently been proposed for related tasks [47], [48] such as edge-to-image, gray-scale-to-color image, day-to-night (or vice versa) photo translation, etc. It is worth noting that, in all these applications, the input and output (target) images are closely related and contain the same contextual information, albeit with a certain degree of corruption. In [34], this problem has become more challenging where each image is transformed to an entirely different image. Moreover, it was also tested whether or not a single network can learn to transform an input image to an output image and vice versa. Evidently, such an inverse problem was the hardest problem tackled in this study due to the distinct and complex patterns and texture of input and output images. To further intricate the problem, the same network was trained to (learn to) transform 4 (target) images from 4 input images. In this study, we adopted the same experimental settings in [34], and accordingly we repeat the experiment 10 times (folds) using the close-up "face" images most of which are taken from the FDDB face detection dataset [49].

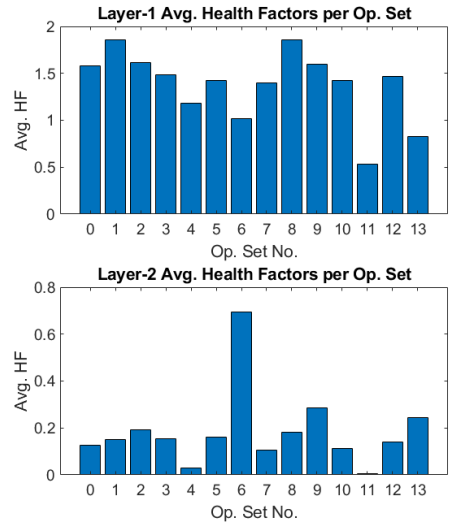

**Figure 16:** Average $HF_\theta^l$ vs. $\theta$ plots for $l=1$ (top) and $l=2$ (bottom) from 10-fold X-validation runs for Image Transformation.

Figure 16 shows the average HF bar plots per operator set from each prior BP run of the 10-fold cross validation. For layer-1, the top three operator sets are usually $\theta = 1\ or\ 8$ and $\theta = 2\ or\ 9$, and $\theta = 0$. It appears that the activation operator does not make a difference in the top-2 operator sets with nodal operators *cubic* and *sine*. The third operator set, $\theta = 0$, corresponds to linear convolution and in 4 out of 10 cross-validations, it was one of the top-3 operator sets. Occasionally, $\theta = 3$, and $\theta = 5$, too with nodal operators *exponential* and *chirp*, respectively, were among the top-3 too. For layer-2, the set $\theta = 6$, with nodal operator *chirp* was always the top operator set in all folds without any exception. It is indeed the most dominant set almost in all runs with the highest final HF usually more than 3 times higher than any other. The 2nd and 3rd top operators were usually $\theta = 9\ or\ 2$ and $\theta = 13$ which correspond to *sine* and again *chirp* nodal operators. Similar to the image syntheses problem, the worst two operators for this layer are $\theta = 4$ *or* $11$ and $\theta = 0$ *or* $7$, (*sinh* and *linear convolution*), respectively. These operators were never found to be amongst the top-5 operators in any fold.

Figure 17 shows the SNR plots of the best CNNs and ONNs among the 10 BP runs for each fold whereas Table 9 presents the average SNR and MSE levels, respectively. Similar arguments and conclusions can be drawn as before. As in the Image Synthesis problem, the elite ONN with $S$=1 surpassed the elite ONN with $S$=3. This is, once again, an expected outcome due to the superiority of the operator set, $\theta = 6$, where the usage of other two sets in the elite ONN with $S$=3 causes a performance loss. Once again, we can see that when there is such a dominant operator set in a hidden layer and limited number of neurons in a compact network, the usage of the best operator set yields the top performance.

On the other hand, the performance of the worst ONN with $S$=3 is better than the worst ONN with $S$=1 and even the CNN. Especially for 2 folds out of 10, SNR > 9dB which indicates that one of the operator set is under-evaluated. This is perhaps because S=30 SPM sessions or M=80 iterations for a session were not sufficiently long enough to capture the true synaptic plasticity level of this operator set. In either case, this shows that for this particular problem, the heterogenous ONN with three different operator sets at each hidden layer significantly surpasses a homogenous network like CNN, which performs only better than the worst ONN with $S$=1.

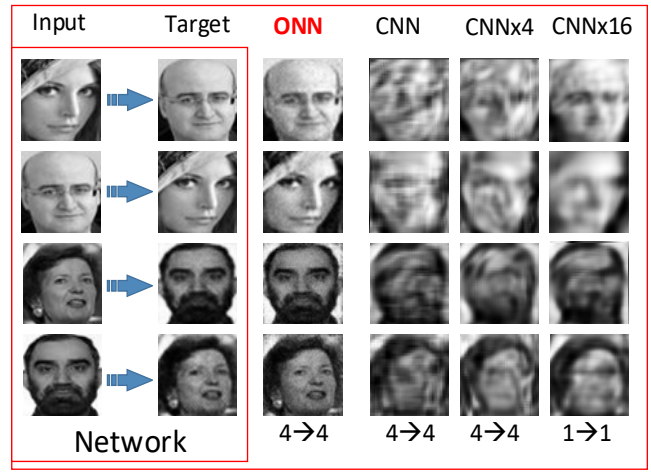

Figure 18: Image transformation of the 1st fold including two inverse problems (left) and the outputs of the elite ONN and three CNN configurations (equivalent, CNNx4 and CNNx16 with 4 and 16 times more parameters, respectively). On the bottom, the numbers of input → target images are shown.

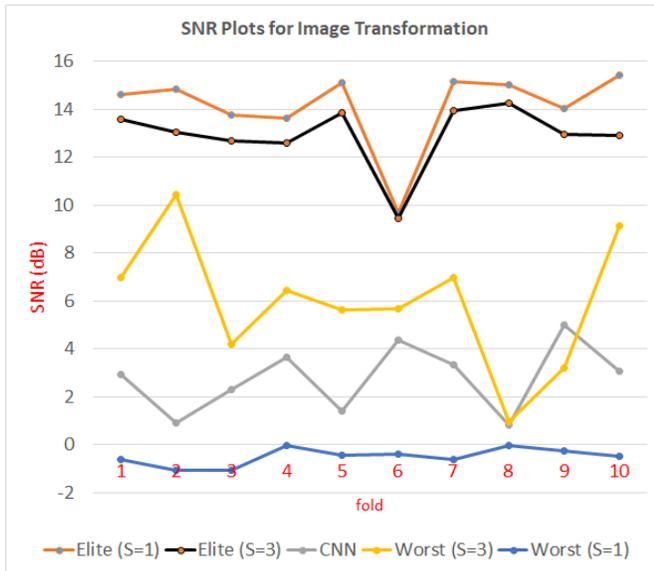

Figure 17: Best SNR levels for each synthesis fold achieved by the top-performing CNNs (grey) and ONNs.

Table 9: Average SNR and MSE achieved in Image Transformation during 10-fold cross-validation. The best (bold) and worst (red) SNR values are highlighted.

| Avg. | Elite (S=1) | Elite (S=3) | CNN | Worst (S=3) | Worst (S=1) |
|---|---|---|---|---|---|
| SNR | **14.07** | 12.86 | 2.78 | 5.86 | -0.49 |
| MSE | **0.9** | 1.19 | 11.94 | 7.45 | 24.33 |

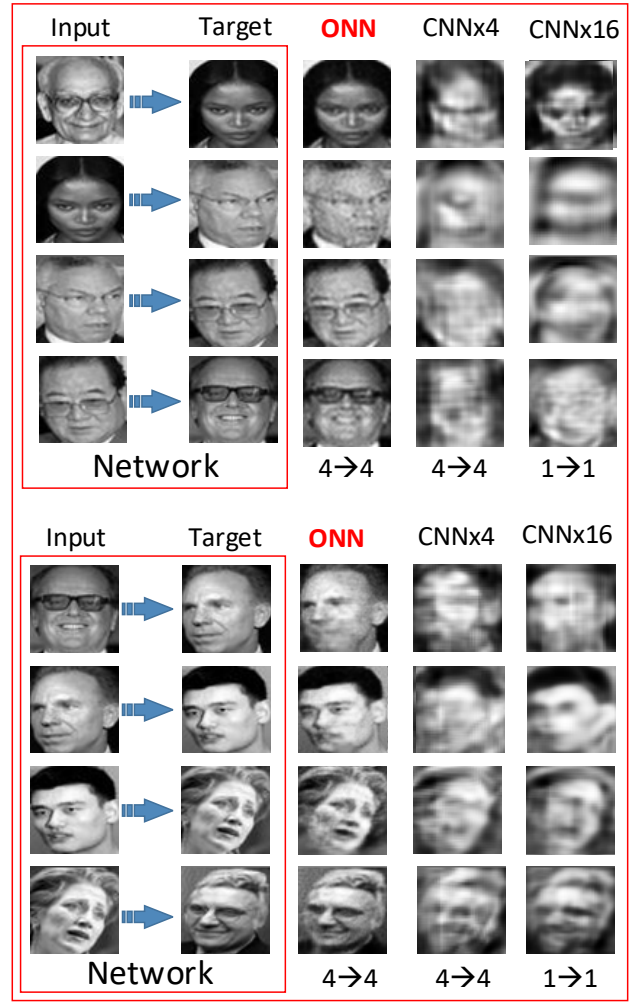

Figure 19: Image transformations of the 3rd (top) and 4th (bottom) folds and the outputs of the elite ONN and three CNN configurations (equivalent, CNNx4 and CNNx16 with 4 and 16 times more parameters, respectively). On the bottom, the numbers of input → target images are shown.

Figure 18 and Figure 19 show the image transformation results for the 1st, 3rd and 4th folds, respectively. In Figure 18, the elite ONN with top-1 operator set has shown a superior performance compared to the three CNN configurations, CNN (default configuration that is same as ONN), CNNx4 and CNNx16 with 4 and 16 times more network parameters, respectively. This is the toughest fold (1st fold) where the two inverse problems (two input and output images are swapped). Although the CNN with 16 times more parameters (having 48 hidden neurons at each hidden layer) is trained over a single (1→1) image transformation, the result is improved but still far from a satisfactory level. Similar observations can be made on the results shown in Figure 19.

As in Image Syntheses problem, the elite (top-1) ONN with 24 neurons also surpassed the ONN (with 48 neurons) configured by GIS in [34] by more than 3 dB SNR on the average. As before, the proposed SPM has resulted better operator sets than the GIS. For instance, for fold 1, it was reported in [34] the best operator sets found by GIS are 0 and 13 for the 1st and 2nd hidden layers whereas the corresponding top operator sets are 1 and 6 by SPM, respectively. With these operator sets, the SNR achieved by the GIS optimized ONN and the elite (top-1) ONN are 10.99dB versus 14.83dB yielding a gap more than 3.5dB. Once again, this shows how crucial to find the best operator set(s) for each hidden neuron to achieve an utmost learning performance.

## V. Conclusions

Synaptic plasticity is a natural process that enables the learning of a new ability, concept or response to changes in the environmental stimuli. This study uses this essential paradigm to configure a highly heterogenous neural network model, the ONN, that is inspired from two basic phenomenon: 1) varying synaptic connections of heterogeneous, non-linear neurons in bio-neurological systems such as the mammalian visual system , 2) direct relation between diversity of neural operators and computational power [4] in biological neural networks wherein adding more neural diversity allows the network size and total connections to be reduced [11]. Empirically, these studies have proven that only the heterogeneous networks with the right operator set and proper training can truly provide the required kernel transformation to discriminate the classes of a given problem, or to approximate the underlying complex function. In neuro-biology, this fact has been revealed as the "neuro-diversity" or more precisely, "the bio-chemical diversity of the synaptic connections" [4]-[9].

To find the right operator set for each hidden neuron, this study proposes to use the synaptic plasticity paradigm periodically during the learning (the prior BP training) process. During the later SPM sessions, those operator sets that exhibit a high synaptic plasticity level or the so-called health factor (HF) are favored whilst the others are suppressed to improve the maturity of the network and finally, they are all ranked based on their HFs. An elite ONN is then formed by using the top-S ranked operator sets in the neurons of each hidden layer. By assigning two different S values, we then exploit the network heterogeneity over the learning performance of the ONNs. To establish a complete "Proof of Concept", the bottom-S ranked operator sets are also used in the so-called "worst" ONNs and evaluated against the elite ONNs, and the CNN. Naturally, it can be expected that the elite ONNs should surpass the native CNN while the worst ONNs should perform the poorest of all. Over the challenging learning problems that are tackled in this study, this expectation holds in general; only in some minority cases, the need for more SPM sessions is observed. This is due to the stochastic nature of backpropagation where the true synaptic plasticity potential of an operator set may not be revealed unless sufficient number of evaluation sessions is performed. A surprising observation worth mentioning is that the conventional CNNs with the same hyper-parameters and configuration may perform even poorer than the worst-3 ONNs (ONNs configured by the bottom-3 operator sets). We foresee that the main reason is due to the homogenous nature of CNNs where its sole operator, the linear convolution, performs rather poor in at least one of the hidden layers. So, the lack of divergence may actually cause it performing even poorer than the worst-3 ONN but not the worst-1 ONN, which also suffers from the limited heterogeneity (i.e., same operator set is used for each layer).

In all problems tackled in this study, the elite ONNs exhibit a superior learning capability conventional CNNs while the performance gap widens when the severity of the problem increases. For instance, in image denoising, the gap between the average SNR levels in train partition was higher than 1.5dB. On a harder problem, image synthesis, the gap widens to above 2.5dB. Finally, on the hardest problem among all, image transformation, the gap exceeded beyond 10dB unless a more complex CNN is used. This is true for the GIS configured ONNs in [34] where SPM has shown to be a better method for searching the top-1 operator set than the GIS. Finally, besides the learning performance, it is worth mentioning that the ONNs also exhibit a superior generalization ability in the test partition of the Image Denoising problem. This is actually an expected outcome since the native operator of the CNN, linear convolution, is an inferior choice especially for layer-2 and it is revealed that this problem requires a higher level of network heterogeneity for the utmost performance. We can conclude that this requires a deeper investigation especially for larger and more complex networks and we believe that the heterogeneity is the key factor for many large-scale machine learning problems and datasets. This will be the topic of our future research.

**Conflict of Interest:** The authors declare that they have no conflict of interest or personal relationships that could have appeared to influence the work reported in this paper